\documentclass[letterpaper, 10 pt, conference]{ieeeconf}
\IEEEoverridecommandlockouts
\pdfoutput=1

\usepackage{graphics}
\usepackage{amsmath}
\usepackage{amsfonts}
\usepackage{graphicx}
\usepackage{xspace}
\usepackage[dvipsnames]{xcolor}
\usepackage[]{subcaption}
\let\labelindent\relax
\usepackage{enumitem}
\usepackage{algorithm}
\usepackage{algorithmicx}
\usepackage[noend]{algpseudocode}
\algrenewcommand\algorithmicindent{0.75em}
\usepackage{caption}
\usepackage{comment}
\usepackage{url}

\newcommand{\ignore}[1]{}

 \def\C{\mathcal{C}} 
  
 \def\I{\mathcal{I}} \def\E{\mathcal{E}}
\def\S{\mathcal{S}} \def\G{\mathcal{G}} 
\def\I{\mathcal{I}}  
 \def\V{\mathcal{V}} 
 \def\W{\mathcal{W}}

\def\dR{\mathbb{R}}

\def\eps{\varepsilon}

\newcommand{\Cpp}{C\raise.08ex\hbox{\tt ++}\xspace}

\newcommand\algname[1]{\textsf{#1}\xspace}

\newcommand\rrt{\algname{RRT}}
\newcommand\rrg{\algname{RRG}}
\newcommand\rrtot{\algname{RRTOT}}

\newcommand\iris{\algname{IRIS}}

\newcommand\irisi{\algname{IRIS\_C}}
\newcommand\irisl{\algname{IRIS\_L}}
\newcommand\irisil{\algname{IRIS\_CL}}
\newcommand\irisile{\algname{IRIS\_CLI}}

\newcommand\ap{\algname{AP}}
\newcommand\pap{\algname{PAP}}

\newcommand\pp{\algname{PP}}
\newcommand\pps{{{\pp}s}\xspace}

\colorlet{pink}{red!40}
\colorlet{light_blue}{cyan!60}

\newcommand{\ep}{{\ensuremath{\eps,p}}\xspace}

\title{\LARGE \bf
Computationally-Efficient Roadmap-based Inspection Planning via Incremental Lazy Search
}

\author{Mengyu Fu$^{1}$,%
\thanks{This project was supported 
by the United States National Science Foundation (NSF) by award 2008475, 
by the US-Israel Binational Science Foundation (BSF) by award 1018193, 
and by the Isaeli Ministry of Science \& Technology (MOST) by award 102583.}%
\thanks{$^{1}$M. Fu and R. Alterovitz are with the Department of Computer Science, University of North Carolina at Chapel Hill, Chapel Hill, NC 27599, USA.
        {\tt\small \{mfu,ron\}@cs.unc.edu}}
Oren Salzman$^{2}$,%
\thanks{$^{2}$Oren Salzman is with Computer Science Department, Technion - Israel Institute of Technology, Israel.
        {\tt\small osalzman@cs.technion.ac.il}}
and Ron Alterovitz$^{1}$%
}

\begin{document}

\maketitle
\thispagestyle{empty}
\pagestyle{empty}

\begin{abstract}
The inspection-planning problem calls for computing motions for a robot that allow it to inspect a set of points of interest (POIs) while considering plan quality (e.g., plan length).
This problem has applications across many domains where robots can help with inspection, including infrastructure maintenance, construction, and surgery.
Incremental Random Inspection-roadmap Search (\iris) is an asymptotically-optimal inspection planner that was shown to compute higher-quality inspection plans orders of magnitudes faster than the prior state-of-the-art method. 
In this paper, we significantly accelerate the performance of \iris to broaden its applicability to more challenging real-world applications.
A key computational challenge that \iris faces is effectively searching roadmaps for inspection plans---a procedure that dominates its running time.
In this work, we show 
how to incorporate lazy edge-evaluation techniques into \iris's search algorithm 
and 
how to reuse search efforts when a roadmap undergoes local changes.
These enhancements, which do not compromise \iris's asymptotic optimality, enable us to compute inspection plans much faster than the original \iris.
We apply \iris with the enhancements to simulated bridge inspection and surgical inspection tasks and show that our new algorithm
for some scenarios
can compute similar-quality inspection plans $570\times$ faster than prior work.

\end{abstract}

%%%%%%%%%%%%%%%%%%%%%
% INTRODUCTION
%%%%%%%%%%%%%%%%%%%%%
% !TEX root =  Fu2021_ICRA.tex

\section{Introduction}
\label{sec:intro}

We consider the problem of \emph{inspection planning} where a robot needs to inspect a set of points of interest (POIs) in a given environment with its on-board sensor while optimizing plan cost.
This problem has numerous applications such as
product surface inspections for industrial quality control~\cite{Raffaeli2013_IJIDeM},
structural inspections with unmanned aerial vehicles (UAVs)~\cite{Cheng2008_IROS, Bircher2015_ICRA, Bircher2016_AR, Almadhoun2016_MWSCAS, Almadhoun2018_IROS},
ship-hull inspections~\cite{Hollinger2012_ICRA, Hollinger2013_IJRR, Englot2013_IJRR},
underwater inspections for scientific surveying~\cite{Bingham2010_FR, Johnson2010_FR, Gracias2013_OCEANS, Tivey1997_EOS},
and patient-anatomy inspections in medical-endoscopic procedures for disease diagnosis~\cite{Kuntz2018_ICRA}.

\begin{figure}
    \centering
    \includegraphics[width=0.95\linewidth]{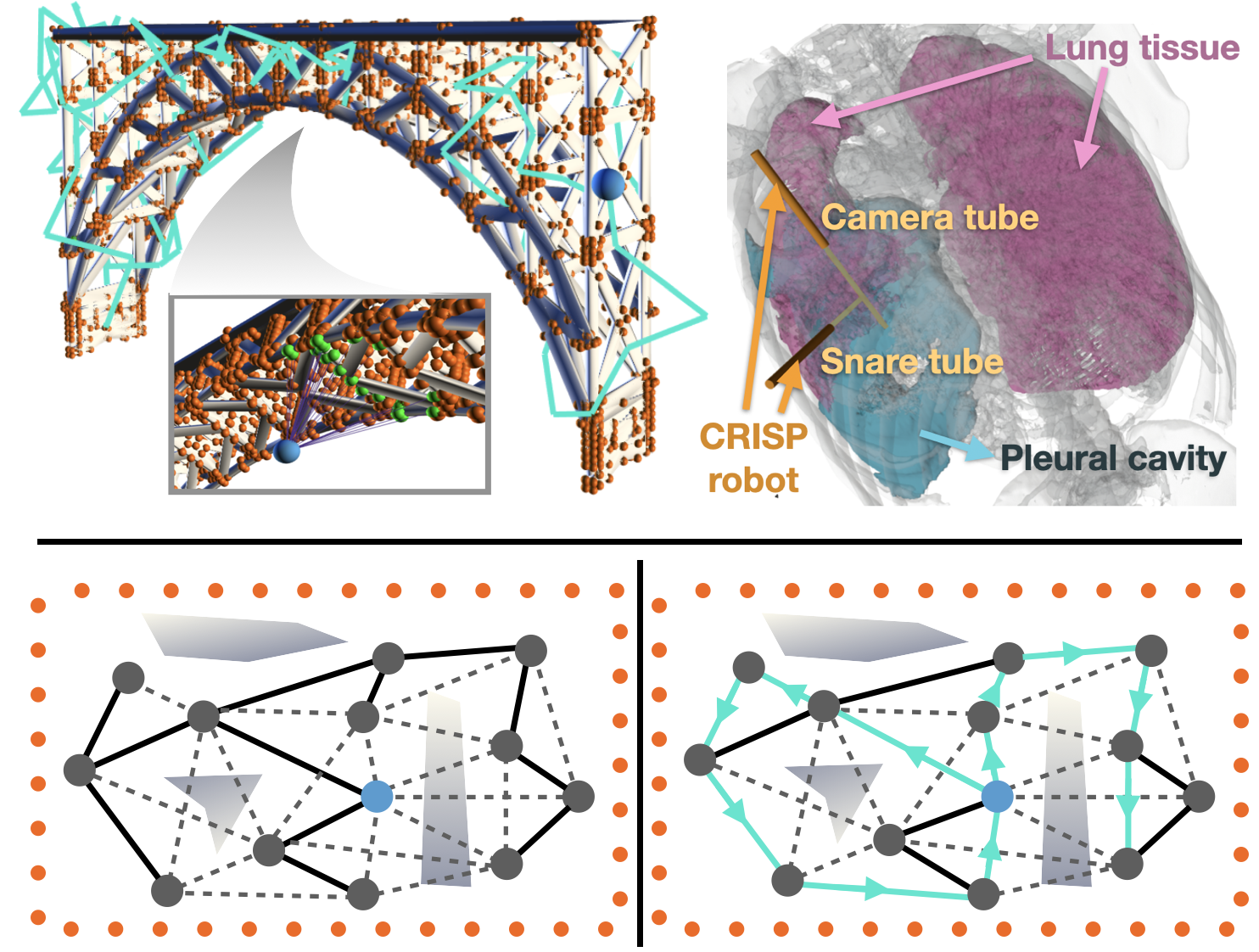}
    \caption{
    Top: Example applications of inspection planning.
    Top left: An unmanned aerial vehicle (blue sphere) inspecting a bridge with its on-board sensor (e.g., camera) by following the inspection plan (aquamarine).
    Points of interest (POIs) are shown as orange spheres (visible POIs are green in zoomed-in part).
    A POI is considered inspected if it is visible to the sensor.
    Top right: 
    The Continuum Reconfigurable  Incisionless Surgical Parallel (CRISP) robot~\cite{Anderson2017_RAL,Mahoney2016_IROS}, a medical robot composed of flexible needle-diameter tubes and equipped with a camera, 
    can perform endoscopic diagnosis in the pleural cavity (the space between the lung surface and chest wall) for a patient with excess fluid surrounding the lung.
    The inspection enables a physician to diagnose underlying disease. 
    Bottom: \iris~\cite{Fu2019_RSS} computes a collision-free plan (aquamarine) to inspect POIs (orange dots).
    It searches a roadmap (dashed edges) implicitly defined by a tree structure (vertices and solid edges) embedded in the configuration space rooted at the start configuration (blue vertex).
    As the roadmap is densified, the resulting inspection plan asymptotically converges to a global optimum.
    }
    \label{fig:cover_fig}
    \vspace{-5mm}
\end{figure}

Roughly speaking, the inspection-planning problem is computationally challenging because we need to simultaneously
reason both about plan cost \emph{and} about inspecting the POIs.
Unfortunately, even computing a minimal-cost plan (without reasoning about inspection) is already PSPACE-hard in the general case~\cite{Halperin2018_Book}.
The problem is exasperated in our setting as the search space over motion plans grows exponentially with the number of POIs to inspect~\cite{Fu2019_RSS}.
Thus, the cost of na\"ively-computed inspection plans may be orders of magnitude higher than the cost of an optimal plan, and computing a high-quality plan (especially with bounds on the solution quality) may be extremely time-consuming.
For time-sensitive applications, generating a high-quality inspection plan in a short computation time (i.e., the time between starting to solve an instance of the problem and getting a result) is critical.
For example, for patient anatomy inspection planned according to pre-operative medical images (Fig.~\ref{fig:cover_fig} top right), a short computation time reduces overall procedure time and makes faster diagnosis possible, which has the potential to improve patient outcomes.
Even for non-time-sensitive applications (e.g., bridge inspection as shown in Fig.~\ref{fig:cover_fig} top left), better computing efficiency means, with the same computing power, we can achieve similar-quality results faster or achieve better-quality results given the same computation time.

To enable faster computation of high-quality inspection plans, we propose algorithmic enhancements to accelerate the performance of Incremental Random Inspection-roadmap Search (\iris)~\cite{Fu2019_RSS}. 
\iris constructs an incrementally-densified roadmap, a graph representing robot states and transitions between the states. 
Then \iris searches over the roadmap for a near-optimal inspection plan. 
In this paper, we accelerate \iris's performance while retaining its formal guarantee of asymptotic optimality.

We accelerate IRIS’s performance in two significant ways.
First, we need to build roadmaps that are compact and yet informative for inspection planning. 
In this paper, we present a simple-yet-effective solution where we reduce the number of samples that fail to inspect previously-uncovered POIs (i.e., POIs not seen by the robot’s sensor from any prior robot state on the roadmap) with coverage-informed sampling 
during roadmap construction. 
Second, we observe that graph search dominates computation time when the graph grows larger, motivating the need for a fast method to effectively search the gradually-densiﬁed roadmap. 
The original \iris runs every search iteration from scratch, disregarding that the roadmap is incrementally densified, but search efforts from previous iterations are potentially helpful in later iterations. 
So in this paper, we show (i) how to incorporate reﬁned lazy edge-evaluation techniques~\cite{CPL14} into IRIS and (ii) how to reuse search efforts when a roadmap locally changes.

These enhancements, which do not compromise \iris's asymptotic optimality, enable us to compute inspection plans much faster than the original \iris.
We evaluate our algorithm in simulation in bridge inspection and surgical inspection scenarios.
Experimental results show that \iris with the enhancements for some scenarios can compute inspection plans of similar-quality $570\times$ faster than the original \iris.

%%%%%%%%%%%%%%%%%%%%%%%%%%%%
% RELATED WORK
%%%%%%%%%%%%%%%%%%%%%%%%%%%%
% !TEX root =  Fu2021_ICRA.tex

\section{Related Work}
\label{sec:related_work}

Inspection planning typically calls for solving two subtasks:
(i)~viewpoints planning that determines a set of viewpoints collectively covering all POIs and 
(ii)~trajectory planning that determines a trajectory connecting the viewpoints.
Many methods solve these subtasks separately.
For viewpoints planning, early methods compute a minimal set of viewpoints by solving the Art Gallery Problem~\cite{Danner2000_ICRA}. 
However, a minimal set of viewpoints doesn't guarantee the optimality of the final plan~\cite{Papadopoulos2013_ICRA}.
So later methods find only a set of viewpoints that satisfies the inspection requirements.
Then the trajectory-planning task is usually formulated using variants of the Traveling Salesman Problem (TSP)~\cite{Edelkamp2016_RAL,Gentilini2013_IROS, Jang2017_ICCAS, Jing2017_IROS}.
To improve the quality of the solution, some use trajectory optimization~\cite{Englot2012_ICAPS, Bogaerts2018_RAL} while others resample viewpoints~\cite{Bircher2015_ICRA, Bircher2016_AR}, or adaptively sample viewpoints~\cite{Almadhoun2018_IROS}.
Unfortunately, this decomposition into two separate steps forgoes any guarantees on the quality of the solution.

Most existing methods for inspection planning (as mentioned above) do not provide formal guarantees on the quality of final solutions.
But recent approaches~\cite{Fu2019_RSS,Papadopoulos2013_ICRA, Bircher2017_Robotica, Kafka2016_ICRA} provide \emph{asymptotic} guarantees by making use of advances in sampling-based motion planners~\cite{Lavalle1998}. 
Specifically, these methods are based on \emph{asymptotically-optimal} motion planners~\cite{gammell-a} such as the probabilistic roadmap* (\algname{PRM*}), the rapidly-exploring random tree* (\algname{RRT*}), and the rapidly-exploring random graph (\algname{RRG})~\cite{Karaman2011_IJRR}. 
Roughly speaking, these methods guarantee that as the number of samples used by the algorithm approaches infinity, the cost of the solution obtained converges to the optimal cost.
Such an asymptotic-optimality guarantee comes at a price of long computation time.
Notable among the inspection planners that do provide formal guarantees on the quality of the solution, \iris was shown to compute higher-quality inspection plans orders of magnitudes faster~\cite{Fu2019_RSS} than the prior state-of-the-art method, Rapidly-exploring Random Tree of Trees (\algname{RRTOT})~\cite{Bircher2017_Robotica}.

For additional related work discussing the connection of inspection planning to other fields, please refer to~\cite{Choset2001_AMAI, Almadhoun2016_IJAR}.

%%%%%%%%%%%%%%%%%%%%%%%
% PROBLEM FORMULATION
%%%%%%%%%%%%%%%%%%%%%%%
% !TEX root =  Fu2021_ICRA.tex

\section{Problem Definition}
\label{sec:pdef}

The robot operates in some physical workspace 
$\W \subset \mathbb{R}^d$ where $d \in \{2,3\}$.
The workspace is cluttered with obstacles~$\W_{\rm obs} \subset \W$.
A robot's configuration $\mathbf{q}$ is a vector of parameters that uniquely defines its state (e.g.,
joint angles for a manipulator arm, pose for an aerial vehicle),
and the set of all configurations~$\C$ is defined as its configuration space or C-space.
Given a configuration~$\mathbf{q}$, we can compute the subset of the workspace~$\W$ occupied by the robot ${\rm Occupancy}(\mathbf{q}) \subset \W$.
We say that $\mathbf{q}$ is collision free if ${\rm Occupancy}(\mathbf{q}) \cap \W_{\rm obs} = \emptyset$ and in-collision otherwise.
This subdivides the C-space $\C$ into the free space $ \C_{\rm free} \subset \C$ and obstacle space $ \C_{\rm obs} = \C \setminus  \C_{\rm free}$.
A robot's path is a mapping $\pi: [0,1] \rightarrow \C$. 
It is \emph{valid} if it follows the kinematic constraints of the robot and if $\forall t \in [0,1], \pi(t) $ is collision free.
In this work we will discretize the path into a finite sequence of configurations $\{\pi(t_0), \ldots, \pi(t_k) \}$ with $k \geq 0$, $t_i < t_{i+1}$, and $t_i \in [0,1]$.
We use linear interpolation between configurations for collision checking.
We are given some cost function ${\rm Cost}: \C \times \C \rightarrow \dR$ and we extend it to paths
${\rm Cost}(\pi) = \sum_{i=0}^{k-1} {\rm Cost}(\pi(t_i), \pi(t_{i+1}))$.
Our framework can deal with general cost functions, but in this particular work, we use path length $\ell(\cdot)$ as cost.

We are given a discrete set $\I  = \{i_0, ..., i_N\} \subset \W$ of \emph{points of interest} (POIs) 
that we need to inspect given some sensor mounted on the robot.
The sensor is modeled by the mapping $\S: \C \rightarrow 2^\I$
(where $2^\I$ is the powerset of $\I$)
that states which POIs can be seen from a given configuration.
A configuration will also be referred to as a \textit{viewpoint}.
We will say that a POI $i \in \I$ is \textit{covered} by a configuration~$\mathbf{q}$ (or that $\mathbf{q}$ \textit{covers}~$i$) if $i \in \S(\mathbf{q})$.
By a slight abuse of notation, we extend the sensor model to paths and have that~$\S(\pi) = \bigcup_{i=0}^{k}\S(\pi(t_i))$ is the inspection coverage of a path~$\pi$.

Given a C-space $\C$, 
a start configuration~$\mathbf{q}_{\rm s} \in \C_{\rm free}$, 
a set of POIs~$\I$, 
a sensor model~$\S$, and a cost function~${\rm Cost}$,
the optimal inspection plan is a valid path
$\pi^* = {\rm argmin}_{\pi \in \Pi}{\rm Cost}(\pi)$.
Here,~$\Pi$ is the set of paths maximizing the inspection coverage, more formally,~$\Pi = \{\pi \vert \pi={\rm argmax}_{\pi \in \Pi_{\mathbf{q}_s}}\vert\S(\pi)\vert\}$,
where~$|\cdot|$ is the cardinality
and 
$\Pi_{\mathbf{q}_s}$ is the set of paths starting from~$\mathbf{q}_{\rm s}$.

%%%%%%%%%%%%%%%%%%%%%%%%%%%%
% METHOD
%%%%%%%%%%%%%%%%%%%%%%%%%%%%
% !TEX root =  Fu2021_ICRA.tex

\section{Algorithmic Background}
\label{sec:background}

\subsection{Incremental Random Inspection-roadmap Search (\iris)}
\label{subsec:iris_intro}

\iris incrementally constructs a sequence of increasingly-dense graphs, or roadmaps, embedded in the C-space and computes an inspection plan over the roadmaps as they are constructed~\cite{Fu2019_RSS}. 
To build the roadmap, \iris builds an \algname{RRG}. 
However, since not all the edges will be used and edge evaluation can be computationally expensive, \iris takes a lazy edge-evaluation approach. This is done by explicitly constructing an  \algname{RRT}~\cite{Lavalle1998} (which is a subgraph of an \algname{RRG} when constructed using the same set of vertices) and leaving all other edges un-evaluated that implicitly define the \algname{RRG}.
The rest of the edges are evaluated on demand during the graph search (for more details see Sec.~\ref{sec:method}).

Computing an optimal-inspection plan on a roadmap~$\G = (\V, \E)$ with~$n$ vertices is computationally hard because we need to compute the shortest path on a so-called \textit{inspection graph}.
Here, each vertex corresponds to a vertex $v \in \V$ of the original graph $\G$ and a subset $I$ of  $\I$, representing a path ending at $v$ while inspecting $I$.
Thus, the inspection graph has $O(n\times \vert2^{\I}\vert)$ vertices.
To search this inspection graph,  \iris employs a novel search algorithm that approximates $\pi^*$, the optimal inspection path on the graph. 
To this end, \iris uses two parameters, $\eps$ and $p$, to ensure that the path obtained from the graph-search phase is no longer than $1+\eps$ the length of $\pi^*$ and covers at least $p$ percent of the POIs covered by~$\pi^*$.
To ensure convergence to the optimal inspection path, $\eps$ is reduced and $p$ is increased between search iterations. This is referred to as \emph{tightening} the approximation factors.

We now briefly describe the graph-search algorithm as it is key to understanding the algorithmic contributions of this work.
For an in-depth description of the search algorithm, the rest of the framework, and its theoretical guarantees, see~\cite{Fu2019_RSS}. 
The search runs an \algname{A*}-like search~\cite{Hart1968_TSSC} but instead of having each node%
\footnote{A node, representing a search state, is different from a vertex, representing a configuration in the underlying roadmap.}
represent a path from the start vertex $\mathbf{q}_s$, each node is a \emph{path pair} \pp that is composed of a so-called \textit{achievable path} (\ap) and \textit{potentially achievable path} (\pap).
As its name suggests, an \ap represents a path in the graph from $\mathbf{q}_s$.
In contrast, a \pap is not necessarily realizable and is used to bound the quality of paths represented by a specific \pp.
A $\pp = (P, \tilde{P})$, where $P$ is the \ap and $\tilde{P}$ is the \pap, is said to be \ep-bounded if 
(i) $\ell(P) \leq (1+\eps)\ell(\tilde{P})$
and
(ii)~$\vert\S(P)\vert \geq p\cdot\vert\S(\tilde{P})\vert$.

Similar to \algname{A*}, the algorithm uses an OPEN and CLOSED list to track nodes that have not and have been considered, respectively.
It starts with the trivial path pair, $\pp_{\mathbf{q}_s}$, where both \ap and \pap represent the trivial path $\{\mathbf{q}_s\}$.

At each iteration, it pops a node from the OPEN list, and checks if the search can terminate. If not, the node is \emph{extended} and added to the CLOSED list. 
While doing so, the algorithm tests if successor nodes can \emph{subsume} or be \emph{subsumed} by another node. 
These two core operations (extending and subsuming) are key to the efficiency of the algorithm and we now elaborate on them for a given path pair $\pp_u = (P_u, \tilde{P_u})$ (here $P_u$ is \ap and $\tilde{P_u}$ is \pap):
\begin{enumerate}[label=(\roman*)]
    \item \textbf{Extending operation:} 
    Extending $\pp_u$ by edge $e = (u,v) \in \E$ 
    (denoted $\pp_u + e$) 
    will extend both $P_u$ and $\tilde{P_u}$ by edge $e$. 
    The resulting path pair is $\pp_v = (P_v, \tilde{P_v})$, where 
    the \ap satisfies 
    $\S(P_v) = \S(P_u) \cup \S(v), \ell(P_v) = \ell(P_u) + \ell(e)$,
    and the \pap satisfies 
    $\S(\tilde{P_v}) = \S(\tilde{P_u}) \cup \S(v), \ell(\tilde{P_v}) = \ell(\tilde{P_u}) + \ell(e)$.
    \item \textbf{Subsuming operation:} 
    Given two path pairs $\pp_1= (P_1, \tilde{P_1})$ and $\pp_2 = (P_2, \tilde{P_2})$ that end at the same vertex, 
    the operation of $\pp_1$ subsuming $\pp_2$ 
    (denoted $\pp_1 \oplus \pp_2$) 
    will create a new path pair 
    $\pp_1\oplus\pp_2 := \left(P_1, \left(\S(\tilde{P_1}) \cup \S(\tilde{P_2}), \min(\ell(\tilde{P_1}), \ell(\tilde{P_2}))\right)\right)$.
\end{enumerate}
Roughly speaking, the subsuming operation allows to reduce the number of paths considered by the search.
However, to ensure that the solution returned by the algorithm approximates the optimal inspection path, subsuming only occurs when the resultant \pp is \ep-bounded.

\subsection{\iris limitations}
\label{subsec:limitations}

\iris was shown to dramatically outperform the prior state-of-the-art in asymptotically-optimal inspection planning~\cite{Fu2019_RSS}.
However, a close analysis of the algorithm's building blocks allows us to pinpoint the computational challenges faced by \iris which, in turn, will allow us to suggest new algorithmic enhancements.
\begin{enumerate}[label=\textbf{(\roman*)}]
    \item \textbf{Configuration sampling in roadmap generation.}
    To generate a roadmap, \iris randomly samples configurations in the C-space.
    Thus, the coverage of a newly-sampled configuration is not accounted for when generating the roadmap.
    This has the potential to increase the roadmap size without adding new (or adding very little) POIs which, in turn, will induce long search times.

    \item \textbf{Edge evaluation during graph search.}
    As mentioned in~\cite{Fu2019_RSS}, the edge-evaluation scheme employed by \iris assumes that edge-evaluation dominates the algorithm's running time. 
    This is true when the size of the roadmap is small but as the number of vertices grows, search dominates the algorithm's running time.

    \item \textbf{Iterative graph search.}
    \iris runs a new graph search after adding configurations to the roadmap.
    In the original formulation, the search tree obtained from previous search episodes is discarded and the search algorithm is run from scratch.
    This can be highly inefficient as often the new graph is very similar to the one used in the previous iterations.
\end{enumerate}
% !TEX root =  Fu2021_ICRA.tex

\section{Method}
\label{sec:method}

We propose several enhancements to \iris corresponding to the algorithmic challenges detailed in Sec.~\ref{subsec:limitations}.%
\footnote{For complete pseudo-code describing our enhancements, see \ref{sec:appendix}.}

% !TEX root =  Fu2021_ICRA.tex

\subsection{Coverage-informed sampling}
\label{subsec:sampling}

To improve the computational efficiency of the original \iris, we propose to replace uniform sampling of configurations (used to construct the roadmap) with an approach that we call 
\emph{coverage-informed sampling} 
that accounts for the inspection coverage of sampled configurations.
Our approach, which bears resemblance to advanced sampling schemes used in motion planning (see, e.g.,~\cite{GBS18,YiTGSS18}), works as follows:
given a randomly-sampled configuration~$\mathbf{q}_{\rm rand}$, 
we accept it directly with some probability~$p_{\rm{accept}} > 0$.
If the configuration was not directly accepted,
we test if  
$
\vert
\S(\V) \cup \S(\mathbf{q_{\rm rand}})
\vert
>
\vert
\S(\V)
\vert,
$
where $\S(\V)$ is the set of POIs collectively covered by roadmap vertices.
If the test passes (namely, if~$\mathbf{q_{\rm rand}}$ covers previously-uncovered POIs), we accept $\mathbf{q_{\rm rand}}$ and add it to the roadmap.
If not, the configuration is rejected.

Randomly accepting $\mathbf{q_{\rm rand}}$ is important for maintaining the theoretical guarantees of \iris (i.e., asymptotic convergence to the optimal inspection path), since $\mathbf{q_{\text{rand}}}$ can be used to reduce path length regardless of its coverage.
This is similar to the way goal-biasing is employed in sampling-based motion planners~\cite{LaValle2006_Book}.

In addition, as graph search dominates the algorithm's running time, we would typically like to initiate a search only when roadmap coverage increases significantly. 
Thus, we define a relaxation parameter $\omega \in (0, 1]$ and run a new search only when 
$|\S(\pi_{\rm prev})| < \omega \cdot p \cdot |\S(\V)|$.
Here, $\pi_{\rm prev}$ is the path returned by the previous search iteration.
This can be seen as a generalization of the original \iris which uses $\omega = 1.0$.
As with the original \iris, after $N_{\max}$ new vertices were sampled without a new search being initiated, we run a new search regardless of the additional coverage.
This allows us to keep reducing the  plan's length even when the coverage does not increase (or increases slowly).
In all our experiments, we use $\omega = 0.9$ and $N_{\max} = 200$.

% !TEX root =  Fu2021_ICRA.tex

\subsection{Refined lazy edge evaluation}
\label{subsec:smart_valid}

Similar to many motion-planning algorithms, the original \iris takes a lazy edge-evaluation approach~\cite{BK00,H15,JSCP15,DS16,SH15a} during graph search.
Specifically, it uses the \algname{LazySP} framework~\cite{DS16} where all edges are assumed to be collision-free and the search is run until a path is found. 
Edges along this path are then evaluated one-by-one. 
If we find an in-collision edge, we discard it and rerun the search. 
If all edges are collision-free then the path is returned.
While this approach was shown to minimize the number of edges evaluated~\cite{HMPSS18}, the computational price of having multiple search episodes can increase overall running times in motion planning algorithms~\cite{MSS18,MCSS19}.

An alternative lazy search algorithm is \algname{LazyA*} which was shown to minimize search efforts~\cite{CPL14}. 
Here, edges are assumed to be collision-free until a node is expanded and only then its incoming edge is evaluated (for additional details, see~\cite{CPL14}).
Unfortunately, we cannot use a \algname{LazyA*}-like approach as-is because of the way paths are subsumed in the near-optimal search algorithm.
Specifically, in our setting, when a node (\pp) is to be extended, it may have already subsumed other nodes (\pps). 
If we discard such a node because its incoming edge is in-collision (as would have been done in \algname{LazyA*}) then we discard all the information about the subsumed nodes, including potentially-valid ones.
This subtlety requires us to additionally validate a node's incoming edges when it is to subsume other nodes.

\begin{figure}
    \centering
    \includegraphics[width=0.95\linewidth]{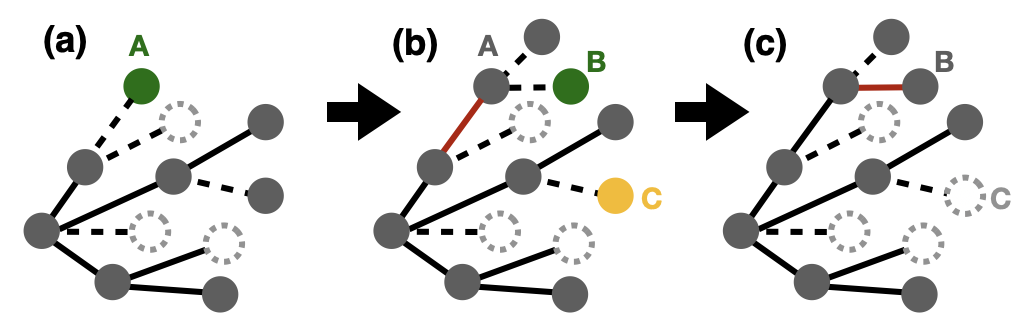}
    \caption{Two different edge evaluation cases.
    Each node in the search tree is a \pp, and nodes with dashed boundaries have been subsumed.
    Solid and dashed lines are validated and yet-to-be validated incoming edges, respectively.
    Red edges are the edges being validated in the current step.
    First case: validating incoming edge of a T-node ((a) to (b)).
    When extending a T-node A, we validate its incoming edge.
    Second case: validating incoming edge when a T-node is to subsume another node ((b) to (c)).
    When a T-node B is to subsume another node C, we validate its incoming edge.
    }
    \label{fig:smart_valid}
    \vspace{-5mm}
\end{figure}

To this end, we consider two types of nodes (\pps)---\emph{trivial}~(T) and \emph{non-trivial}~(NT) which correspond to \pps that did not and did subsume another \pp, respectively.
Note that by our definition a T-node may be a descendent of an NT-node.
We validate the incoming edge of a T-node when either
(i)~it is extracted from the OPEN list (similar to \algname{LazyA*})
or when
(ii)~it is to subsume another node (thus becoming an NT-node which implies that all incoming edges of NT-nodes are validated).
For a visualization, see Fig.~\ref{fig:smart_valid}.

% !TEX root =  Fu2021_ICRA.tex

\subsection{Incremental search by reusing efforts across iterations}
\label{subsec:reuse}

In the original \iris, every new search is run from scratch and the only information shared between search iterations is edge validity.
Fortunately, reusing search information across different search iterations is a well-studied problem~\cite{FMN00,RR96,KLF04} that has been successfully used in many motion-planning algorithms (see, e.g.,~\cite{H15,SH15a,MSS18,SH16, Gammell2015_ICRA}). 

In typical approaches, between two  search episodes, all nodes and data structures (e.g., OPEN and CLOSED lists in \algname{A*}) are stored.
After an edge or vertex is added or removed from the graph, we first identify all \emph{inconsistent} nodes (namely, whose cost changed due to the change in the graph), then their cost is updated and they are re-added to the relevant data structures.
Unfortunately, we cannot use these methods as-is because of the way  the approximation factor changes between search iterations.

The crux of the problem is that even if there is no change in the graph, tightening the approximation factors $\eps$ and $p$ may cause nodes (i.e., path pairs) in the search tree to no longer be \ep-bounded (for the new values of $\eps$ and $p$).
To this end, we first discuss how we ensure that before executing a new search, all nodes in the search tree (namely, in the OPEN and CLOSED lists) are \ep-bounded when there are no changes to the graph. 
We then describe how to account for the additional vertices and edges added to~$\G$.

\subsubsection{Accounting for tightening of the approximation factors}
\label{sec:tight}

Recall that in our search algorithm, we have two basic operations on nodes: extending the associated path pair \pp by an edge and subsuming another path pair.
The former can only decrease the (relative) gap between the achievable path (\ap) and the potentially achievable path (\pap) of the \pp.
In contrast, the latter may increase the (relative) gap between the \ap and \pap of the \pp.
Thus, after tightening the approximation factors $\eps$ and $p$, nodes that are still \ep-bounded may have unbounded predecessors.

If a node is no longer \ep-bounded, we need to ``rollback`` previous subsuming operations to obtain nodes that are \ep-bounded. 
This requires us to store for each \pp not only the \ap and \pap but also a list of all \pps that were subsumed and may dramatically increase the program's memory footprint.
To reduce unnecessary memory usage, we perform the following optimization: when the subsuming operation $\pp_1 \oplus \pp_2$ is performed, if
$\S(P_1) \supseteq \S(\tilde{P_2}), \ell(P_1) \leq \ell(\tilde{P_2})$, then $\pp_2$ is not stored in the subsumed list.
This is because the \pap of $\pp_2$ (which is a lower bound on the solution that can be obtained using that node) is strictly dominated by the \ap of $\pp_1$ (which, by definition, can be obtained).
To further reduce memory usage, instead of storing $\pp_2$ directly, we store the predecessor of $\pp_2$. 
Since the ending vertex of $\pp_2$ is the same as $\pp_1$ thus is known, we can easily reconstruct $\pp_2$ from its predecessor if need to.
Since $\pp_2$ is not directly stored, the nodes subsumed by $\pp_2$ (if any) are inherited by $\pp_1$ when $\pp_1$ subsumes $\pp_2$.

\begin{figure}
    \centering
    \includegraphics[width=0.95\linewidth]{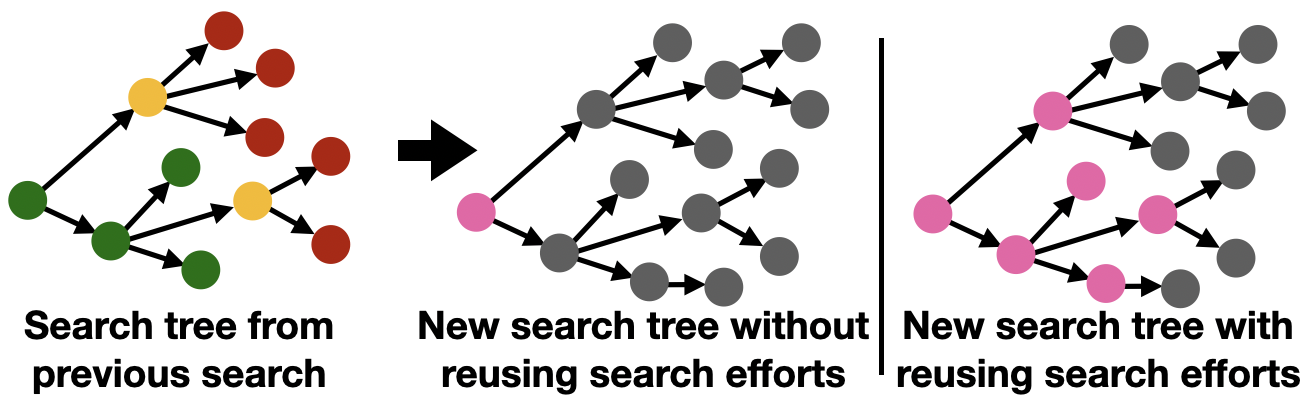}
    \caption{Visualization of how we reuse search efforts. 
    Reusable nodes are in green, boundary nodes are in yellow and non-reusable nodes are in red.
    Pink nodes are obtained directly from previous search while gray nodes are newly constructed in the current search.
    }
    \label{fig:node_reuse}
    \vspace{-5mm}
\end{figure}

To reuse search efforts, we define two characteristics of nodes.
First, a node is \textit{revealed} (if stored in the OPEN or CLOSED list) or \textit{hidden} (if subsumed and stored in the subsumed list of another node).
And a node can be a \textit{reusable}, \textit{boundary}, or \textit{non-reusable} node
(see also, Fig.~\ref{fig:node_reuse}):
\begin{enumerate}[label=\textbf{(\roman*)}]
    \item \textbf{Reusable nodes}---a node is reusable if it is \ep-bounded and all its predecessors are also \ep-bounded.

    \item \textbf{Boundary nodes}---a node is a boundary node if it is
no longer \ep-bounded, but all its predecessors are.
    
    \item \textbf{Non-reusable nodes}---a node is non-reusable if it has
at least one predecessor that is no longer \ep-bounded.
\end{enumerate}
As the name suggests, reusable nodes can be used as-is. 
Thus, we keep revealed reusable nodes in the OPEN or CLOSED list, depending on where they were when the previous search terminated.

Before we discuss how we handle boundary and non-reusable nodes, 
notice that given a boundary node $\pp_v$ with parent $\pp_u$, we have that extending $\pp_u$ by the edge $(u,v)$ 
(denoted as $\pp_u + (u,v)$) 
yields a new path pair $\hat{\pp}_v$ to $v$ that is \ep-bounded (with an empty list of subsumed nodes).
Thus, we start by adding all revealed boundary and non-reusable nodes into a list which we call RELEASE and removing them from the OPEN or CLOSED list.
Nodes from RELEASE are popped one by one until it is empty. 
For each node, $\pp_v$, popped from RELEASE we
(i)~add $\pp_v$ (if it is a reusable node) or $\pp_u +(u,v)$ (if it is a boundary node) to the OPEN list (while testing for dominance as in the original \iris),
and
(ii)~add all of $\pp_v$'s subsumed nodes to RELEASE in case it is a boundary or a non-reusable node.

While a complete proof of the correctness of our approach is beyond the scope of this paper, we notice that:
(i) after the above operations, all revealed nodes (either in the OPEN or CLOSED list) are \ep-bounded;
(ii) for every node $\pp_v$ (either revealed or hidden) from the previous search iteration, if $\pp_v$ is not in the CLOSED list, then either $\pp_v$ or one of its predecessors is either in the OPEN list or in the subsumed list of a node in the OPEN or CLOSED list, which means that all nodes are accounted for.
The above properties guarantee that the new search still results in a near-optimal solution.

\subsubsection{Accounting for roadmap updates}
\label{sec:roadmap-updates}

To further account for new vertices and edges, for any node $\pp_u$ in the CLOSED list corresponding to a vertex $u \in \V$ that has an edge $(u,v)$ to a newly-inserted vertex $v$, we add the new successor $\pp_u + (u,v)$ to the OPEN list.

\subsubsection{Putting it all together}

To summarize, our new search algorithm receives as input not only $\G$, $\mathbf{q}_s$, $\varepsilon$ and $p$, but also the OPEN and CLOSED lists from the previous search iteration (if it is the first iteration then both are empty).
We start by adding the node used to obtain the path in the previous iteration back into the OPEN list (if it is the first iteration, we add $\pp_{\mathbf{q}_s}$).
We then continue to update the OPEN and CLOSED lists to account for the tightening of the approximation factors (Sec.~\ref{sec:tight}) and then we update the OPEN list to account for the roadmap updates (Sec.~\ref{sec:roadmap-updates}).
Finally, we run our search algorithm without any changes.

%%%%%%%%%%%%%%%%%%%%%%%%%%%%
% RESULTS
%%%%%%%%%%%%%%%%%%%%%%%%%%%%
% !TEX root =  Fu2021_ICRA.tex

\section{Results}
\label{sec:results}

\begin{figure*}
    \centering
    \includegraphics[width=\linewidth]{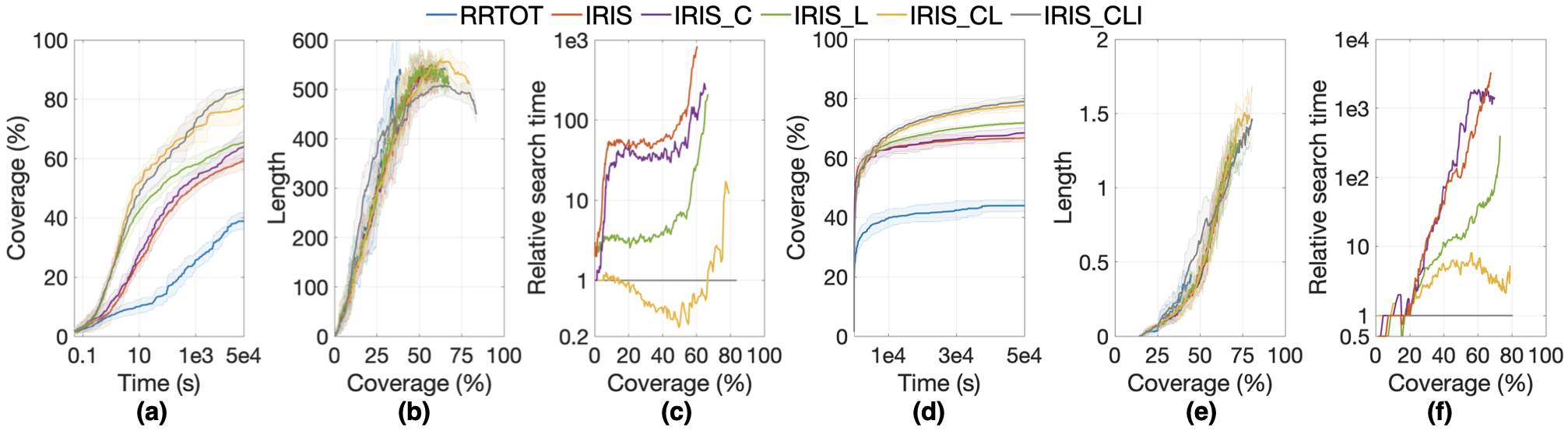}
    \caption{
    Result comparison of different methods.
    (a)--(c): Bridge inspection scenario.
    All variants used hyperparameters
    $p_{\rm accept} = 0.05, p_0 = 0.85, \varepsilon_0 = 10.0$.
    (d)--(f): Pleural cavity inspection scenario.
    All variants used hyperparameters
    $p_{\rm accept} = 0.1, p_0 = 0.9, \varepsilon_0 = 15.0$.
    }
    \label{fig:result}
    \vspace{-6mm}
\end{figure*}

We evaluated our algorithmic enhancements on two simulated scenarios, a bridge-inspection task and a medical-inspection task.
In each scenario, we compared 
the original \iris with four variants, namely
 \irisi, \irisl, \irisil, and \irisile, where 
\algname{C}, \algname{L}, and \algname{I}
stand for
\textbf{C}overage-informed sampling, 
refined \textbf{L}azy edge validation, and 
\textbf{I}ncremental search by reusing search efforts,
respectively.
For reference, we also run \rrtot~\cite{Bircher2017_Robotica}.
Our evaluation metrics include path coverage and path length as a function of the running time, and for \iris and its variants, we also look at the relative search efficiency compared to \irisile. 
We define search efficiency as, for a given inspection coverage~$s$, the relative time each algorithm spends on search (and not on roadmap construction or edge evaluation) to return a path $\pi$ that satisfies $|\S(\pi)| \geq s$ for the first time.
Search efficiency is compared only in Fig.~\ref{fig:result}~(c) and (f), while Fig.~\ref{fig:result}~(a) and (d) use the total running time.

Our implementation is based on the publicly-available \Cpp implementation of \iris~\cite{Fu_GitHub} (new code corresponding to the improvements suggested in this work are also incorporated to this repository).
All experiments were run on a dual 2.1GHz 16-core Intel Xeon Silver 4216 CPU and 100GB of RAM.
All experiments were run for $50,000$ seconds. 
The initial values of \ep are denoted as $\varepsilon_0, p_0$ and we updated them with
$p_i = p_{i-1} + f\cdot(1 - p_{i-1}), \varepsilon_i = \varepsilon_{i-1} + f\cdot(0 - \varepsilon_{i-1})$,
where $f$ is a parameter controlling the tightening of the approximation factors.
We used $f = 10^{-4}$.
Finally, additional results can be found in \ref{sec:additional_results}.

\subsection{Bridge inspection scenario}
\label{subsec:bridge}

Almost $40\%$ of the bridges in the US exceed their 50-year design life~\cite{ASCE2017}, and regular inspections are critical to ensure bridge safety. 
Existing inspection methods are typically expensive, and using UAVs to autonomously inspect bridges could potentially reduce inspection time and costs.
In this scenario, a UAV with a camera is tasked with inspecting a bridge (Fig.~\ref{fig:cover_fig} top left), which includes 3,346 POIs extracted from a 3D mesh model.

The UAV's C-space is $\mathbb{R}^3\times\mathcal{SO}(2)$:
it may translate in 3D and rotate around its vertical axis, and the camera can further rotate around the pitch axis.
The camera is modeled as having a field of view of $94$ degrees and an effective inspecting range of $10$ meters.

Fig.~\ref{fig:result}~(a),~(b) show that all \iris variants have better performance than the original \iris since they achieve better inspection coverage faster while keeping similar plan lengths.
Specifically, \irisile achieves the best performance, reaching $23\%$ more inspection coverage than the original \iris
and
is $570\times$ faster in reaching a coverage of $60\%$.
\irisile also achieves $44\%$ more inspection coverage than \rrtot.
Fig.~\ref{fig:result}~(c) compares the speedup in graph search time of \irisile to reach a given inspection coverage comparing to other variants. 
As we can see, for search efficiency,
\irisile is $830\times$ 
faster than the original \iris for a coverage of $60\%$, 
and it is roughly $12\times$
faster than the second-best method, \irisil, for a coverage of $78\%$.

\subsection{Pleural cavity inspection scenario}
\label{subsec:pleural}

A pleural effusion is a serious medical condition in which excess fluid builds up between the lungs and chest wall and can cause lung collapse. 
Pleural effusions may result from many different diseases, including cancer.
To effectively diagnose the cause of pleural effusion, doctors need to visually inspect the inner surface of the pleural cavity, the fluid-filled space between the lungs and the chest wall.
In this scenario, a CRISP robot~\cite{Mahoney2016_IROS, Anderson2017_RAL} with a chip-tip camera performs the endoscopic diagnosis (Fig.~\ref{fig:cover_fig} top right).
The pleural cavity is segmented from a patient Computerized Tomography (CT) scan, and we densely sample 
4,203
POIs on the surface of the pleural cavity.

Composed of flexible needle-diameter tubes, the CRISP robot requires smaller incisions compared to traditional endoscopic instruments, thus reducing patient pain and shortening the recovery time.
The tubes are inserted into the patient body separately and assembled into a parallel structure with snares.
When manipulating the tubes outside the body, the shape of the robot inside the body changes accordingly, changing the pose of the chip-tip camera to inspect surrounding anatomy.
In our scenario, we use a two-tube CRISP robot.
At the entry point, each tube can rotate in three dimensions (yaw, pitch, and roll) and translate in one dimension (insert or retract).
So the system has a C-space $\C \subseteq \mathcal{SO}(3)^2\times \mathbb{R}^2$.

The robot's forward kinematics (FK) is computationally expensive: 
computing the shape of the robot requires numerical methods to model the torsional and elastic interactions between all the flexible tubes of the parallel structure~\cite{Mahoney2016_IROS}.
Thus FK-involved procedures, including roadmap construction and edge validation during graph search, are time-consuming.
In this scenario, for the initial $50\%$ coverage, the original \iris shows better efficiency since fewer edges are validated.
However, Fig.~\ref{fig:result}~(d),~(e) show that as $\G$ grows, the performance of the original \iris plateaus as the time spent on search begins to dominate, while other variants keep making progress.
\irisile reaches $12\%$ more inspection coverage than the original \iris
and
is $6\times$ faster in reaching a coverage of $67\%$.
\irisile also achieves $35\%$ more inspection coverage than \rrtot.
When it comes to search efficiency, Fig.~\ref{fig:result}~(f) shows \irisile is over $3,200\times$ faster than the original \iris for a coverage of $67\%$ and
$5\times$ faster than \irisil for a coverage of $79\%$.

%%%%%%%%%%%%%%%%%%%%%%%%%%%%
% CONCLUSION
%%%%%%%%%%%%%%%%%%%%%%%%%%%%
% !TEX root =  Fu2021_ICRA.tex

\section{Conclusion \& future work}
\label{sec:discussion}

In this paper, 
we introduced a faster inspection planning algorithm based on enhancements to \iris.
The new algorithm uses coverage-informed sampling to construct more effective roadmaps
and
uses two enhanced search techniques to improve the computational efficiency of the 
near-optimal search algorithm employed by the original \iris. 
More specifically, one is refined lazy edge validation that saves computation time for search when the roadmap is densified,
and the other is reusing search efforts when a roadmap undergoes local changes.
Simulation experiments in two scenarios show the improved search algorithm is dramatically faster than the original one.
With better computational efficiency, we can achieve similar-quality plans with significantly shorter computation time while retaining \iris's asymptotic optimality.

The \iris framework can be further improved.
In this work, the coverage-informed sampling method simply rejects samples to control the size of the roadmap.
This yields good performance, but a more effective way of determining good viewpoints 
would potentially be more beneficial.
A different possible approach to build a compact and informative roadmap is sparsifying the roadmap. Here we draw inspiration from roadmap sparsification techniques for motion planning (see, e.g.,~\cite{SSAH14,DB14}) that demonstrated how to reduce a given roadmap's size with very little compromise on the quality of resulting plan.

\section*{Acknowledgment}

We thank Nadav Elias for his contribution to ideas that led to coverage-informed sampling in this work.

\bibliographystyle{IEEEtran}
\bibliography{IEEEabrv,references-fixed,references,newrefs}

% !TEX root =  Fu2021_ICRA.tex

\clearpage

\setcounter{section}{0}
\renewcommand{\thesection}{Appendix~\Alph{section}}

\setcounter{subsection}{0}
\renewcommand{\thesubsection}{\Alph{subsection}}

\section{Pseudo-code}
\label{sec:appendix}

In this section we provide pseudo-code for the new improvements.
To make this section self-contained, there is some text that repeats descriptions provided in the main body of this work.

\subsection{Original \iris}

Before introducing \irisile, we first briefly review the original \iris algorithm~\cite{Fu2019_RSS}.
Please note for Alg.~\ref{alg:iris} and Alg.~\ref{alg:near-optimal}, \iris is in black and \textcolor{blue}{blue} text, while for \irisile, \textcolor{OliveGreen}{green} lines are added and \textcolor{blue}{blue} lines are modified.

Let~$\C$ be the configuration space, $\mathbf{q}_s$ a start configuration, $\S$ a sensor model, $\I$ a set of POIs, and $ell$ a distance metric, all as defined in Sec.~\ref{sec:pdef}.
In the original \iris algorithm, configuration sampling and near-optimal graph search are performed in an interleaved manner (see Alg.~\ref{alg:iris}).

In the sampling phase, a roadmap $\G$ is constructed as a Rapidly-exploring Random Graph (\rrg), which is implicitly defined by a Raplidly-exploring Random Tree (\rrt).
Denote $\G = (\V, \E)$, where $\V$ is the set of all vertices and $\E$ is the set of all edges.
Each vertex $v \in \V$ is a collision-free configuration~$\mathbf{q} \in \C_{\rm free}$, whose inspection coverage is $\S(\mathbf{q})$, and each edge $e \in \E$ is the motion connecting two configurations, whose length is defined by $\ell(e)$.
Since the roadmap is implicitly defined, only edges in the \rrt is validated during construction.
Then in the search phase, a near-optimal plan is found.
A plan $\pi$ is \textit{near-optimal} for some approximation parameters $p \in [0, 1]$ and $\varepsilon \geq 0$ if 
$|\S(\pi)| \geq p\cdot|\S(\pi^*)|$
and
$\ell(\pi) \leq (1 + \eps) \cdot \ell(\pi^*)$,
where $\pi^*$ is the optimal plan on $\G$ that has the shortest length among all feasible trajectories that covers the  maximum number of POIs.

The near-optimal graph search (Alg.~\ref{alg:iris}, line~\ref{line:search}) is a core component in \iris's framework.
The details are shown in Alg.~\ref{alg:near-optimal}.
Each node in the search is a \textit{path pair} (\pp) composed of an \textit{achievable path} (\ap) and a \textit{potentially achievable path} (\pap).
We denote $\pp_v = (P_v, \tilde{P_v})$, where $\pp_v$ denotes a path pair that starts from $\mathbf{q}_s$ and ends at vertex $v \in \C$, here $P_v$ is an \ap and $\tilde{P_v}$ is a \pap.
For formal definitions of a \pp, an \ap, and a \pap, please refer to~\cite{Fu2019_RSS}.
Roughly speaking, an \ap is an achievable sequence of vertices, by tracing back the predecessors of the \pp, such a sequence can be reconstructed; while for a \pap, there's no way to reconstruct the sequence of vertices and we don't even require that such a path exists. 
A \pap is defined by the pair $\S(\tilde{P_v})$ and $\ell(\tilde{P_v})$ bounding  the potentially-best inspection coverage and plan length can be achieved via all paths encoded in the path pair.
Note that by definition, 
$\ell(\tilde{P_v}) \leq \ell(P_v)$
and
$\S(\tilde{P_v}) \supseteq \S(P_v)$.

\begin{algorithm}[t!]
\caption{Incremental Random Inspection-roadmap Search(IRIS) \\
\textbf{Input:} $\C$, $\mathbf{q}_s$, $\S$, $\I$, $p_0$, $\varepsilon_0$ \\
\textbf{Colors:} \textcolor{blue}{modified}
}

\begin{algorithmic}[1]
\State $\V \leftarrow \{\mathbf{q}_s\}$, $\E \leftarrow \emptyset$, $\G \leftarrow \left(\V, \E\right)$
\State {$\pi \leftarrow$ \{$\mathbf{q}_s$\}}
\State $p \leftarrow p_0$, $\varepsilon \leftarrow \varepsilon_0$
\While{not TIMED\_OUT}
    \State $\G \leftarrow$ \textcolor{blue}{ExpandRoadmap($\C$, $\G$, $\I$)}
    \label{line:sampling}
    \Comment{{\footnotesize Sampling}}
    \State $p, \varepsilon \leftarrow$ UpdateApproximation()
    \label{line:approx}
    
    \If{not \textcolor{blue}{NeedNewSearch($\G$, $\pi$, $p$)}}
    \label{line:new_search}
        \State \textbf{continue}
    \EndIf
    
    \State $\pi \leftarrow$ \textcolor{blue}{NearOptimalGraphSearch($\G$, $\mathbf{q}_s$, $p$, $\varepsilon$)}
    \Comment{{\footnotesize Search}}
    \label{line:search}
\EndWhile

\State \textbf{return} $\pi$

\end{algorithmic}
\label{alg:iris}
\end{algorithm}

We define two operations defined on \pps:
\begin{enumerate}[label=(\roman*)]
    \item \textbf{Extending operation:} extending $\pp_u$ by edge $e = (u,v) \in \E$ will extend both $P_u$ and $\tilde{P_u}$ by edge $e$. The result path pair is $\pp_v = (P_v, \tilde{P_v})$, where 
    \ap satisfies $\S(P_v) = \S(P_u) \cup \S(v), \ell(P_v) = \ell(P_u) + \ell(e)$,
    and \pap satisfies $\S(\tilde{P_v}) = \S(\tilde{P_u}) \cup \S(v), \ell(\tilde{P_v}) = \ell(\tilde{P_u}) + \ell(e)$.
    We denote the extending operation as $\pp_v = \pp_u + e$.
    \item \textbf{Subsuming operation:} Subsuming between two \pps can only happen when both \pps start from the same vertex (this is non-trivial since all \pps starts from $\mathbf{q}_s$) and ends at the same vertex. Assume we have ${\rm PP_1} = (P_1, \tilde{P_1})$ and ${\rm PP_2} = (P_2, \tilde{P_2})$ both ends at vertex $v$, the result path pair of ${\rm PP_1}$ subsuming ${\rm PP_2}$ is defined as $\pp_1 \oplus \pp_2 := \left(P_1, \left(\S(\tilde{P_1}) \cup \S(\tilde{P_2}), \min(\ell(\tilde{P_1}), \ell(\tilde{P_2}))\right)\right)$.
\end{enumerate}
Subsuming is a core operation to achieve near-optimal search.
The search starts with the root node  $\pp_{\mathbf{q}_s} = \left(\{\mathbf{q}_s\}, (\S(\mathbf{q}_s), 0)\right)$ where the \pap and \ap have exactly the same inspection coverage and path length.
When subsuming happens during the search, the \pap gradually increases its inspection coverage and/or decreases path length when compared to the \ap.
To guarantee near-optimality of the result, we ensure that all path pairs are  \ep-bounded.
Here, a \pp is said to be \ep-bounded if its \ap $P$ and \pap $\tilde{P}$ satisfy both
$|\S(P)| \geq p \cdot |\S(\tilde{P})|$
and
$\ell(P) \leq (1 + \varepsilon)\cdot \ell(\tilde{P})$.
By guaranteeing that all \pps in the closed set and open list are \ep-bounded for the given $p$ and $\varepsilon$, Alg.~\ref{alg:near-optimal} finds near-optimal inspection plans (see~\cite{Fu2019_RSS} for a proof sketch).

\begin{algorithm}[t!]
\caption{
NearOptimalGraphSearch\\
\textbf{Input:} $\G, \mathbf{q}_s, p, \varepsilon$ \\
\textbf{Colors:} \textcolor{blue}{modified}, \textcolor{OliveGreen}{added}
}
    \begin{algorithmic}[1]
        \State ${\rm CLOSED}, {\rm OPEN} \leftarrow$ \textcolor{blue}{InitializeLists($\mathbf{q}_s, p, \varepsilon$)}
        \label{line:init}
        \vspace{2mm}
        \While{${\rm{OPEN}} \neq \emptyset$}
            \State $\pp_u \gets$ \rm{OPEN{}.extract\_min}()
            \label{line:pop}
        \vspace{2mm}
            {\textcolor{OliveGreen}{
            \State{\textbf{if} {not Valid($\pp_u)$} \textbf{then}}
            \label{line:pop_check}
            \State{\hspace{0.25cm}\textbf{continue}}
            \label{line:pop_check_2}
            \Comment{{\footnotesize Edge validation}}
            }}
        \vspace{2mm}
            \If{$\S(\tilde{P_u}) == \S(\G.\V)$}
                \Comment{{\footnotesize $\tilde{P_u}$ is the \pap of $\pp_v$}}
                \label{line:goal_}
                \State{\textcolor{OliveGreen}{OPEN.insert($\pp_u$)}}
                \textcolor{OliveGreen}{\Comment{{\footnotesize For search efforts reuse}}}
                \label{line:insert_result}
                \State \Return $P_u$ \Comment{{\footnotesize ${P_u}$ is the \ap of $\pp_u$}}
                \label{line:return}
            \EndIf

        \vspace{2mm}
            \For{$e=(u,v) \in$ {Neighbors}($u, \G$)}
            \label{alg:neighbor}
            \State{$\pp_v \leftarrow \pp_u + e$}
            \Comment{{\footnotesize Extending operation}}
            \label{line:extend}
            \State{\textcolor{blue}{AddNewNode($\pp_v, {\rm CLOSED}, {\rm OPEN}, p, \varepsilon$)}}
            \label{line:add-new-node}
            \EndFor
        \vspace{2mm}
            \State{\rm{CLOSED.insert}}($\pp_u$)
            \label{line:closed}
            
        \EndWhile
        \State \textbf{return} NULL
    \end{algorithmic}
\label{alg:near-optimal}
\end{algorithm}

When a new node $\pp_v$ is generated after some other node was extended (Alg.~\ref{alg:near-optimal}, line~\ref{line:extend}),
we perform the following operations to add the node while ensuring that all \pps are \ep-bounded:
\begin{enumerate}[label=(\roman*)]
    \item For any node $\pp_v' \in {\rm CLOSED}$ in the closed set, if 
    $\S(\tilde{P_v'}) \supseteq \S(\tilde{P_v})$
    and
    $\ell(\tilde{P_v'}) \leq \ell(\tilde{P_v})$,
    then the new node is dominated.
    In such cases we update the node in the closed set with $\pp_v' \leftarrow \pp_v' \oplus \pp_v$.\footnote{Please note, this update doesn't change $\S(\tilde{P_v'})$ nor $\ell(\tilde{P_v'})$, it only keeps record that such subsuming operation happened.}
    \item If a node cannot be dominated by a  node in the closed set, we check if for any  node in the open list $\pp_u' \in {\rm OPEN}$, if
    $\pp_v' \oplus \pp_v$ is \ep-bounded, then the new node is dominated, we update the node in the open list with $\pp_v' \leftarrow \pp_v' \oplus \pp_v$.
    \item Finally if a new node cannot be dominated by any node in either the closed set or the open list, it may still subsume nodes in the open list .
    For any node in the open list $\pp_u' \in {\rm OPEN}$, if $\pp_u \oplus \pp_u'$ is \ep-bounded, $\pp_u'$ is dominated, we remove it from the open list and update $\pp_u$ with $\pp_u \leftarrow \pp_u \oplus \pp_u'$, and have ${\rm OPEN}.{\rm insert}(\pp_u)$.
\end{enumerate}

\subsection{Refined lazy edge evaluation}

Recall that trivial (T) and non-trivial (NT) nodes are referring to \pps that did not and did subsume other \pps, respectively.
Further recall that  we perform edge evaluation when a T-node subsumes another node.
Thus, any node popped from the OPEN list is either a
(i)~T-node;
or an 
(ii)~NT-node that is already checked to be valid.
Thus, we add line~\ref{line:pop_check}-\ref{line:pop_check_2} in Alg.~\ref{alg:near-optimal} to reject invalid T-nodes.
Moreover, Alg.~\ref{alg:near-optimal} line~\ref{line:add-new-node} is also modified, checking the leading edge of a T-node when it is subsuming another node.

\subsection{Incremental search by reusing efforts across iterations}

Since this part of the algorithm is already described in details in Sec.~\ref{subsec:reuse}, here we directly provide the pseudo-code (Alg.~\ref{alg:init_lists},~\ref{alg:release_subsumed}) as a complement.

\begin{algorithm}[t!]
\caption{InitializeLists\\
         \text{Input:} $\mathbf{q}_s, \ep$}
    \begin{algorithmic}[1]
        \If{ initial search}
            \State{${\rm CLOSED} \leftarrow \emptyset$, ${\rm OPEN} \leftarrow \{\pp_{\mathbf{q}_s}\}$}
        \Else
            \State{${\rm CLOSED}, {\rm OPEN} \leftarrow$ RetrieveLatestLists()}
            \label{alg:init_lists:retrieve}
        \EndIf
    \vspace{2mm}
        \For{$\pp_v \in {\rm CLOSED} \cup {\rm OPEN}$}
            \State{MarkNodeType($\pp_v$)}
            \label{alg:init_lists:mark}
        \EndFor
    \vspace{2mm}
    \State {RELEASE $\leftarrow \emptyset$}
    \For{$\pp_v \in {\rm CLOSED} \cup {\rm OPEN}$}
        \If{$\pp_v$ is reusable}
            \State{\textbf{continue}}
        \EndIf
        
        \State{${\rm RELEASE}.{\rm insert}(\pp_v)$}
        \label{line:insert-release}
        
        \If{$\pp_v \in {\rm CLOSED}$}
            \State{${\rm CLOSED}.{\rm erase}(\pp_v)$}
        \EndIf
        
        \If{$\pp_v \in {\rm OPEN}$}
            \State{${\rm OPEN}.{\rm erase}(\pp_v)$}
        \EndIf
    \EndFor

\vspace{2mm}  
    \For{$\pp_v \in {\rm RELEASED}$}
        \State{RecursivelyRelease($\pp_v, {\rm CLOSED}, {\rm OPEN}, \ep$)}
        \label{line:recursive-release}
    \EndFor

\vspace{2mm}
    \For{$\pp_u \in {\rm CLOSED}$}
        \For{$e=(u,v) \in \text{Neighbors}(u, \G)$}
            \If{$v$ is new vertex}
                \State{$\pp_v \leftarrow \pp_u + e$}
                \State{AddNewNode($\pp_v, {\rm CLOSED}, {\rm OPEN}, \ep$)}
                \label{line:new-successor}
            \EndIf
        \EndFor
    \EndFor
    
    \State{\textbf{return} ${\rm CLOSED}, {\rm OPEN}$}
        
    \end{algorithmic}
\label{alg:init_lists}
\end{algorithm}

\begin{algorithm}[t!]
\caption{RecursivelyRelease\\
         \text{Input:} $\pp_v, {\rm CLOSED}, {\rm OPEN}, \ep$}
    \begin{algorithmic}[1]
        \If{$\pp_v$ is reusable}
        \label{alg:release_subsumed:check}
            \State{AddNewNode($\pp_v, {\rm CLOSED}, {\rm OPEN}, \ep$)}
            \label{line:add-reusable}
            \State{\textbf{return}}
        \Else{ \textbf{if} $\pp_v$ is boundary \textbf{then}}
            \State{$\pp_v \leftarrow \pp_v.{\rm predecessor()} + e$}
            \State{AddNewNode($\pp_v, {\rm CLOSED}, {\rm OPEN}, \ep$)}
            \label{line:add-boundary}
        \EndIf
        \For{$\pp_v' \in \pp_v$.subsumed\_list()}
            \State {RecursivelyRelease($\pp_v', {\rm CLOSED}, {\rm OPEN}, \ep$)}
            \label{line:recursive}
        \EndFor
    \end{algorithmic}
\label{alg:release_subsumed}
\end{algorithm}

\section{Additional Results}
\label{sec:additional_results}

Here we provide a decomposition of running time of \iris and \irisile.

\subsection{Bridge inspection scenario}

\begin{figure*}[tbh]
\begin{subfigure}{0.5\linewidth}
    \centering
     {\includegraphics[width=\linewidth]{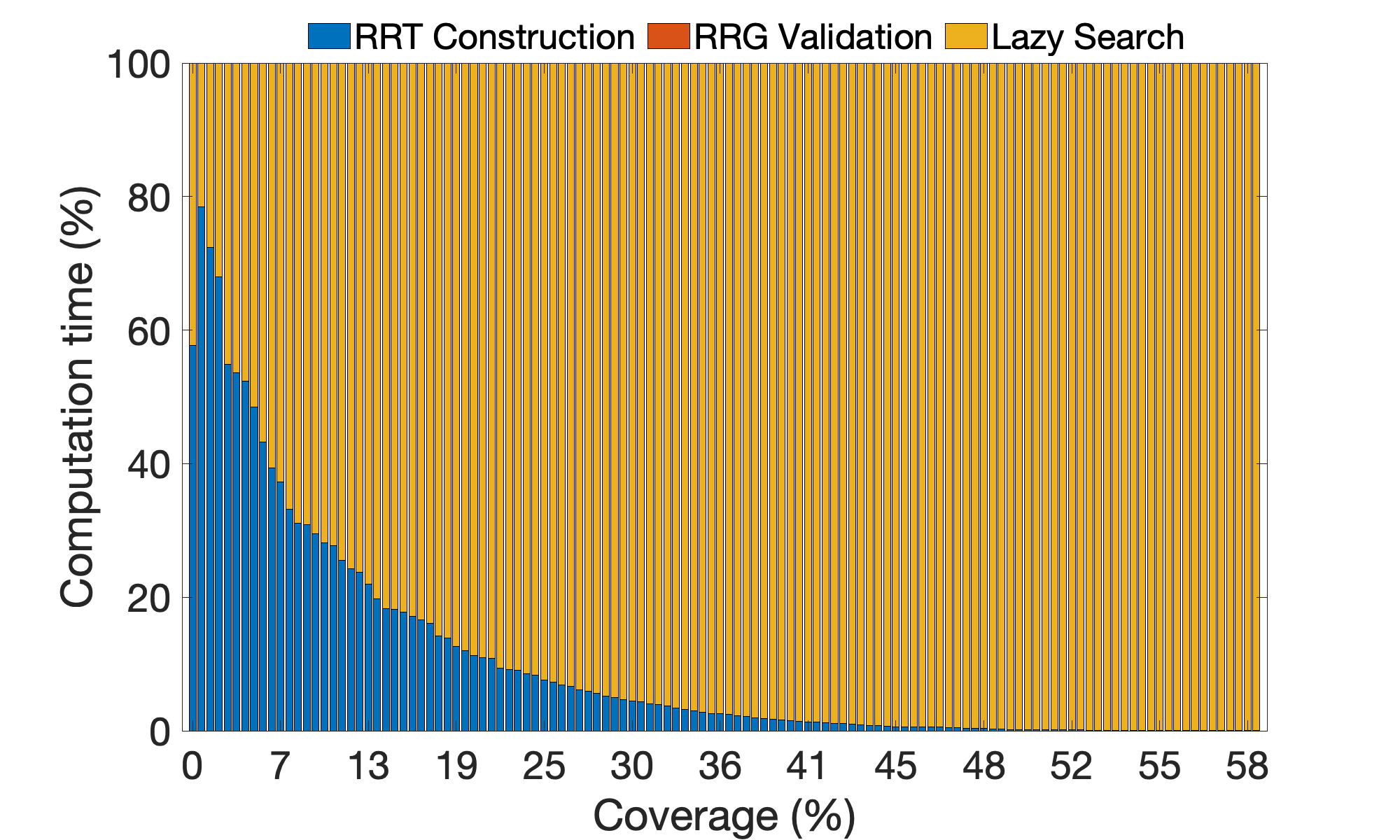}}
  \caption{}
  \label{fig:iris-drone}
\end{subfigure}
\begin{subfigure}{.5\linewidth}
    \centering
     {\includegraphics[width=\linewidth]{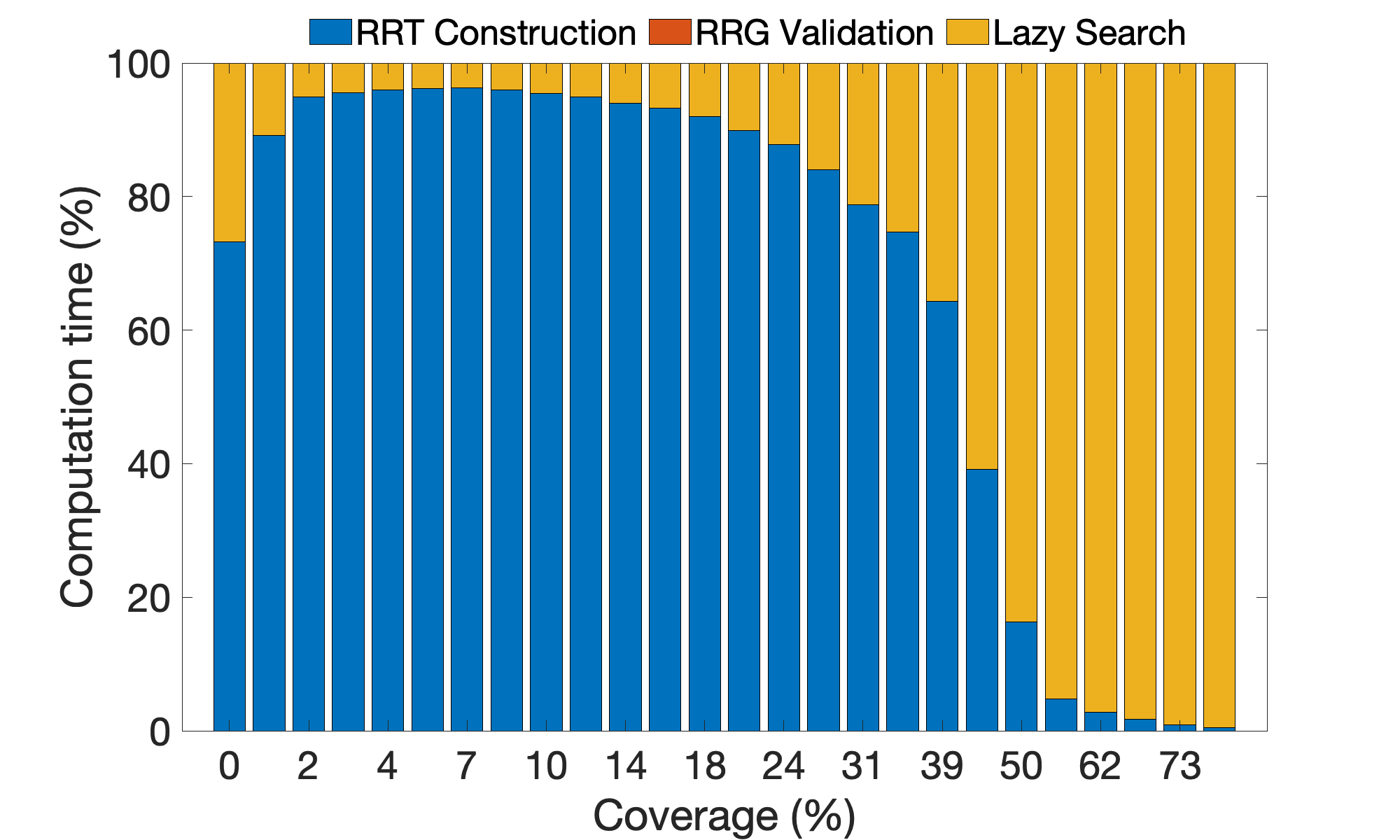}}
  \caption{}
  \label{fig:irisile-drone}
\end{subfigure}
\begin{subfigure}{0.5\linewidth}
    \centering
     {\includegraphics[width=\linewidth]{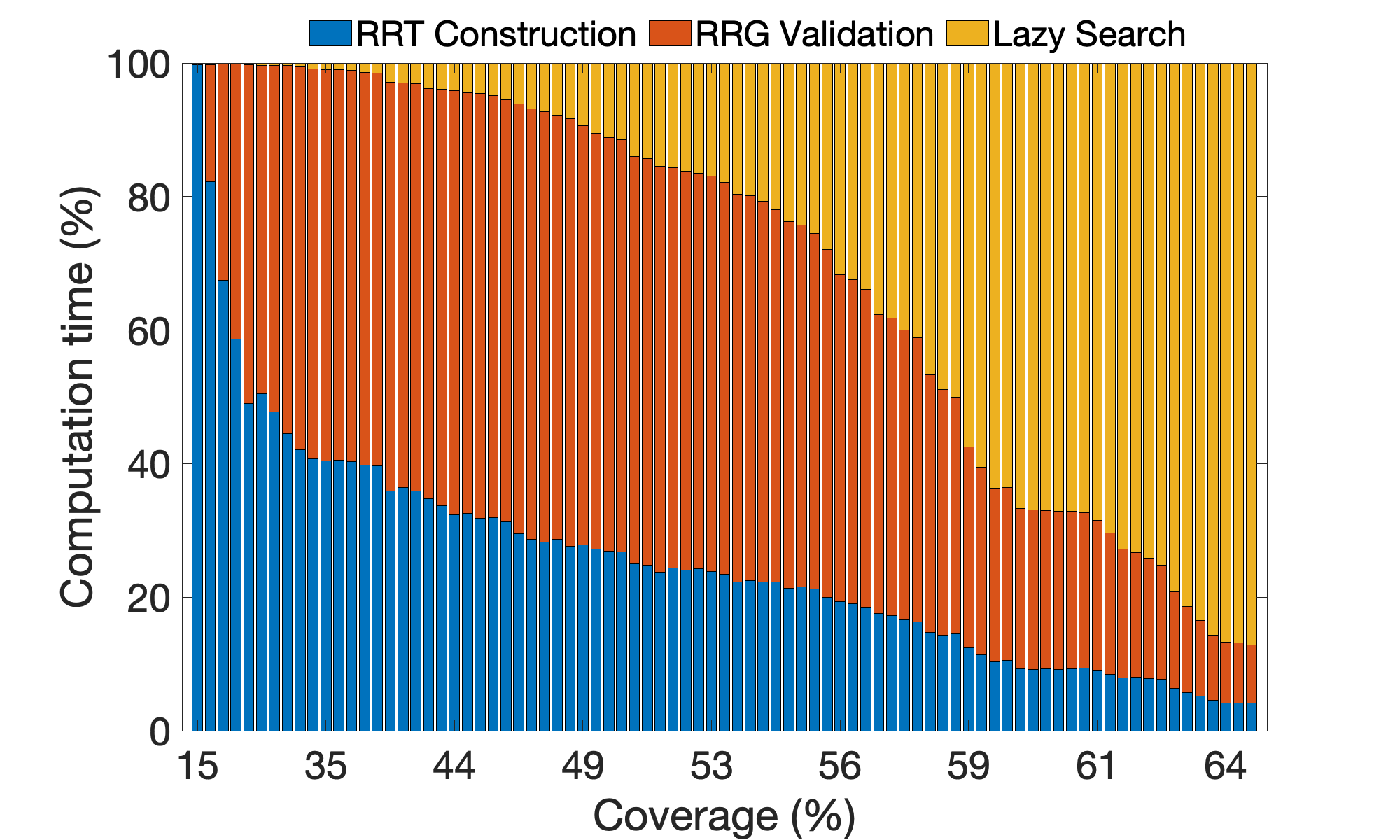}}
  \caption{}
  \label{fig:iris-crisp}
\end{subfigure}
\begin{subfigure}{.5\linewidth}
    \centering
     {\includegraphics[width=\linewidth]{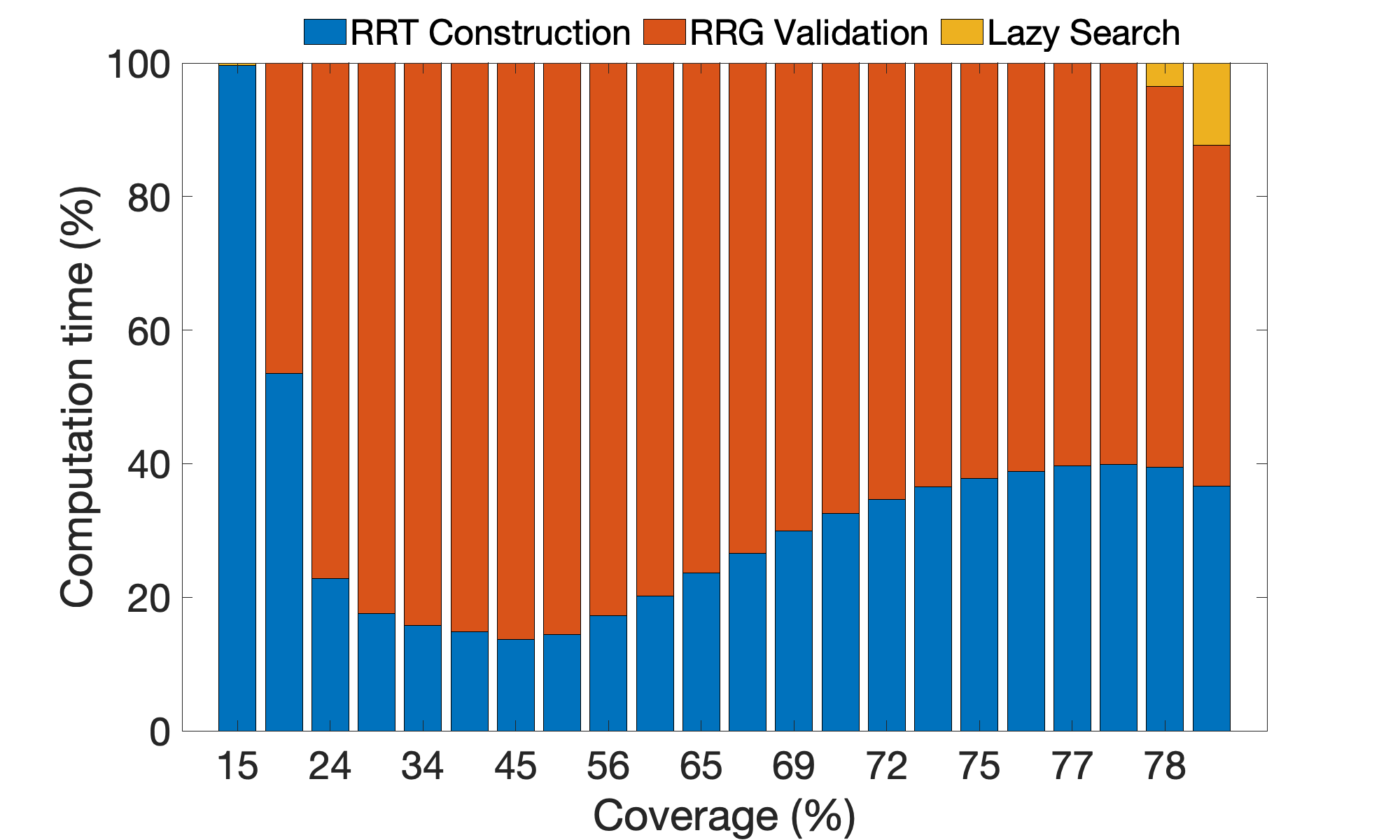}}
  \caption{}
  \label{fig:irisile-crisp}
\end{subfigure}
  \caption{
  Running time decomposition.
  \protect (\subref{fig:iris-drone})~\iris for bridge inspection scenario.
  \protect (\subref{fig:irisile-drone})~\irisile for bridge inspection scenario.
  \protect (\subref{fig:iris-crisp})~\iris for pleural cavity inspection scenario.
  \protect (\subref{fig:irisile-crisp})~\irisile for pleural cavity inspection scenario.
  }
 \label{fig:time-decomp}
\end{figure*}

The decomposition of the running time  of \iris and \irisile are shown in Fig.~\ref{fig:iris-drone} and Fig.~\ref{fig:irisile-drone}, respectively.

For the bridge-inspection scenario, edge validation is computationally cheap since the we do interpolation between configurations without considering the dynamics of the robot.
From the charts, the time spent on edge validation is almost negligible compared to the time spent on roadmap construction and search.
Additionally, note that with coverage-informed sampling and using the relaxation parameter $\omega$ (introduced in Sec.~\ref{subsec:sampling}), we need to run significantly fewer iterations to reach the same inspection coverage.

\subsection{Pleural cavity inspection scenario}

The decomposition of the running time  of \iris and \irisile are shown in Fig.~\ref{fig:iris-crisp} and Fig.~\ref{fig:irisile-crisp}, respectively.

For the pleural-cavity inspection scenario, edge validation is computationally expensive since the FK of the crisp robot is hard to compute.
We can see from the charts that lazy search started to dominate the computation time in the original \iris while in \irisile, the time spent on search is relatively low compared to roadmap construction and edge validation.
Similar to the bridge-inspection scenario, with coverage-informed sampling and the relaxation parameter $\omega$ (introduced in Sec.~\ref{subsec:sampling}), we need to run significantly fewer iterations to reach the same inspection coverage.

\end{document}